\documentclass[runningheads]{llncs}
\usepackage{graphicx}
\usepackage{amssymb} 
\usepackage{hyperref}
\usepackage{booktabs}
\usepackage{multirow}

\begin{document}
\title{
Do Not Sleep on Traditional Machine Learning 
\thanks{\url{https://idioms.thefreedictionary.com/sleep+on+(someone+or+something)}}
}
\subtitle{Simple and Interpretable Techniques Are Competitive to Deep Learning for Sleep Scoring}
%
%
\newcommand*\samethanks[1][\value{footnote}]{\footnotemark[#1]}

\author{
Jeroen Van Der Donckt \orcidID{0000-0002-9620-888X} \thanks{contributed equally} \and
Jonas Van Der Donckt \orcidID{0000-0003-1440-7410} \samethanks \and
Emiel Deprost \orcidID{0000-0001-5323-5506} \and
Nicolas Vandenbussche \orcidID{0000-0001-8400-1163} \and
Michael Rademaker \orcidID{0000-0001-7124-692X} \and 
Gilles Vandewiele \orcidID{0000-0001-9531-0623} \and 
Sofie Van Hoecke \orcidID{0000-0002-7865-6793}
}
\authorrunning{J. Van Der Donckt et al.}
%
\institute{
IDLab, Ghent University - imec, Technologiepark 126, 9052 Zwijnaarde, Belgium
}
\maketitle              
\begin{abstract}
Over the last few years, research in automatic sleep scoring has mainly focused on developing increasingly complex deep learning architectures. 
However, recently these approaches achieved only marginal improvements, often at the expense of requiring more data and more expensive training procedures.
Despite all these efforts and their satisfactory performance, automatic sleep staging solutions are not widely adopted in a clinical context yet.
We argue that most deep learning solutions for sleep scoring are limited in their real-world applicability as they are hard to train, deploy, and reproduce. Moreover, these solutions lack interpretability and transparency, which are often key to increase adoption rates. 
In this work, we revisit the problem of sleep stage classification using classical machine learning. Results show that competitive performance can be achieved with a conventional machine learning pipeline consisting of preprocessing, feature extraction, and a simple machine learning model. 
In particular, we analyze the performance of a linear model and a non-linear (gradient boosting) model.
Our approach surpasses state-of-the-art (that uses the same data) on two public datasets: \texttt{Sleep-EDF SC-20} (MF1 0.810) and \texttt{Sleep-EDF ST} (MF1 0.795), while achieving competitive results on \texttt{Sleep-EDF SC-78} (MF1 0.775) and \texttt{MASS SS3} (MF1 0.817). %
We show that, for the sleep stage scoring task, the expressiveness of an engineered feature vector is on par with the internally learned representations of deep learning models.
This observation opens the door to clinical adoption, as a representative feature vector allows to leverage both the interpretability and successful track record of traditional machine learning models.

\end{abstract}
%
%

\section{Introduction}
Humans spend around one-third of their lives asleep. Sleep is the single most effective process to reset our brain and body health each day, making it fundamental to human health~\cite{besedovsky2019sleep,cappuccio2017sleep}.
At the same time, sleep disorders are a common and increasingly prevalent public health issue~\cite{ohayon2011epidemiological}, e.g., in the U.S. millions of people's lives are impaired by sleep disorders~\cite{national2011national}. This makes objective, quantitative diagnosis of sleep quality and associated disorders a major topic in medicine and research~\cite{wulff2010sleep}. 

As part of clinical sleep analysis, physicians collect and analyze polysomnography (\textsc{PSG}) data. PSG is the continuous monitoring of brain activity with electroencephalography (\textsc{EEG}), eye movements with electrooculography (\textsc{EOG}), muscle activity with electromyography (\textsc{EMG}), heart rhythm with electrocardiography (\textsc{ECG}), and respiration. Typically, such a recording is performed during inpatient overnight at the hospital while the patient sleeps. A \textsc{PSG} is the current gold standard for objective assessment of sleep continuity and quantify associated pathologies, including sleep-related breathing disorders, narcolepsy and sleep-related movement disorders~\cite{penzel2000computer}.

Sleep scoring, also known as ``sleep staging" or ``sleep stage classification", consists of the classification of 30-seconds periods of \textsc{PSG} (called ``epochs") in different sleep stages. 
Sleep stage classification is one of the fundamental technical investigations at the basis of clinical decision-making in sleep medicine diagnostics and treatment efficacy evaluation~\cite{shrivastava2014interpret,wulff2010sleep}.
This task is done visually by a sleep expert. Sleep experts mostly rely on guidelines (e.g., proposed by the American Academy of Sleep Medicine (AASM)~\cite{berry2012rules}) for the determination of sleep stages. Each epoch is classified into one of the following five sleep stages: wakefulness~\textsc{w}, stage~\textsc{n1}, stage~\textsc{n2}, stage~\textsc{n3}, and rapid eye movement~(\textsc{rem}). The manual annotation of the recorded data is a complex and time-intensive process which takes a well-trained physician up to two hours to score one whole hypnogram of about 8 hours of sleep~\cite{malhotra2013performance}. Moreover, sleep stage classification is prone to subjective bias, resulting in lower than desired inter-scorer and intra-scorer agreement (i.e., ~83\% inter-scorer agreement~\cite{rosenberg2013american}, and ~90\% intra-scorer agreement~\cite{fiorillo2019automated}). In particular, \textsc{n1} and \textsc{n3} have inter-scorer agreements as low as 63 and 67\% respectively, which raises questions on their usefulness~\cite{stephansen2018neural}.
This disagreement stems from (i) sleep stages being a discretization of a continuous process~\cite{berthomier2020exploring}, (ii) visual scoring inherently being limited by how the human visual and cognitive systems interpret the data (the latter includes individual differences in scoring experience), and (iii) how the PSG data is presented to the annotator~\cite{silber2007visual}.
Although numerous solutions have been devised in an effort to automate sleep stage classification, to date, no system has completely replaced humans as the gold standard~\cite{fiorillo2021deepsleepnetlite}. Among many reasons, the most prominent ones are (i) the limited adoption rate of machine learning (and in particular deep learning) in the healthcare domain (e.g., aversion to technology, usability or technical limitations)~\cite{fichman2011editorial}, and (ii) security and privacy issues as often powerful resources from the cloud are required to perform deep learning-based scoring~\cite{fiorillo2019automated}.

Today, literature on automated sleep scoring research is mainly concerned with deep learning solutions. We observe a general trend of applying increasingly complex deep learning solutions, resulting in only marginal gains. Moreover, these small improvements often require more data and more expensive training procedures.
Yet, clinical acceptance of deep learning solutions is often directly hindered by the obstacles that deep learning introduces, such as being hard to train, deploy, and reproduce, while lacking interpretability~\cite{fiorillo2019automated,amann2020explainability,fichman2011editorial}. As a result, the latest research focuses on bringing data-efficiency or interpretability to deep learning models.
However, as opposed to trying to solve the disadvantages associated with deep learning techniques post-hoc, we believe that one should instead use techniques that do not introduce these issues in the first place~\cite{rudin2019stop}.

Therefore, in this paper, we challenge the concept that deep learning paradigms are necessary to develop performant data driven models for automatic sleep stage classification. 
To do so, we employ a conventional machine learning pipeline, consisting of three steps: preprocessing, feature extraction, and modeling. We compare the performances of both a linear and a non-linear model.

The contributions of this work are twofold;
\begin{enumerate}
    \item We propose a novel approach for automatic sleep scoring that performs on par with deep learning solutions and is easy to reproduce. We open source our results and pipelines at \url{https://github.com/predict-idlab/sleep-linear}.
    \item We discuss the impact of our results and show that traditional machine learning deserves more attention within the automatic sleep stage classification domain. In particular, this work opens the door to clinical acceptance of similar simple pipelines.
\end{enumerate}

The remainder of this paper is as follows. In Section~\ref{sec:related_work} we discuss existing research. Section~\ref{sec:approach} describes our proposed pipeline. In Section~\ref{sec:experiments}, we present the results of our solution of four datasets. Section~\ref{sec:discussion} discusses why we achieve satisfying performance with simple models and what this could imply for future research. Finally, we end with a conclusion in Section~\ref{sec:conclusion}.

\section{Related work}\label{sec:related_work}
In this section, we will describe related research on automatic sleep scoring. First, we will zoom in on previous work that uses classical machine learning, followed by the more recent advancements that employ deep learning.

\subsection{Sleep scoring using classical machine learning}
Earlier work in this domain applied conventional machine learning pipelines consisting of preprocessing, feature extraction, and modeling. We observe several similarities among these works. Most approaches used single channel EEG as input~\cite{li2017hyclasss,koley2012ensemble,hassan2017decision,alickovic2018ensemble,khalighi2013automatic,liang2012automatic}, and extracted multi-domain (e.g., temporal and spectral) features~\cite{li2017hyclasss,koley2012ensemble,lajnef2015learning,hassan2017decision,khalighi2013automatic,malafeev2018automatic}. In general, non-linear classification models were applied, such as random forests~\cite{li2017hyclasss,malafeev2018automatic} and support vector machines~\cite{koley2012ensemble,lajnef2015learning,khalighi2013automatic,alickovic2018ensemble}.

We believe that in many cases the contributions of the aforementioned research is limited as (i) small datasets were used (less than 30 subjects), (ii) limited features were extracted (less than 40), and (iii) no effort has been made to allow reproduction of the results as either the work was performed on proprietary data, or the code was not made publicly available.

However, there are some exceptions. Li et al.~\cite{li2017hyclasss} evaluated their proposed approach on an open dataset of 198 subjects, showing that a conventional machine learning pipeline scales to larger datasets. Khalighi et al.~\cite{khalighi2013automatic} indicated (after extensive feature extraction) that multimodal EEG, EOG, and EMG channels as input results in the best performance.  The work of Malafeev et al.~\cite{malafeev2018automatic} compared traditional machine learning (random forest on 20 features) to deep neural networks. They concluded that deep neural networks are superior in their generalization ability, however the performance difference was not really pronounced.  

Given the limitations of the above contributions, comparing these research efforts on classical machine learning to state-of-the-art deep learning solutions is hard to realize. Only the recent work of Vallat et al.~\cite{vallat2021open} on classical machine learning for sleep scoring is comparable to recent deep learning research. In their paper, the authors aimed at making a broadly applicable sleep scoring algorithm available\footnote{RobustSleepNet~\cite{guillot2021robustsleepnet} and U-Sleep~\cite{perslev2021usleep} are deep learning solutions with the same (highly relevant) goal. These works focussed on developing a solution that is robust to (i) arbitrary PSG montages (and protocols), and (ii) various clinical populations.}. 
The proposed approach employs a tree-based gradient boosting model together with post-processing smoothing. Results indicated that their classical machine learning solution performs 2-4\% lower than state-of-the-art deep learning approaches\footnote{We cannot facilitate direct comparison with the work of Vallat et al.~\cite{vallat2021open} as their solution is evaluated on other datasets. However, we do compare directly with the deep learning approaches that are referenced in their comparisons.}. This work builds further on the above observations, i.e., investigating the performance of conventional pipelines.

\subsection{The use of deep learning and clinical acceptance}\label{sec:related_work_DL}
In recent years, research focus for the sleep scoring task has mainly been concentrated towards deep learning algorithms. Among the utilized deep learning architectures are auto-encoders~\cite{tsinalis2016automatic1}, fully connected neural networks~\cite{dong2017mixed}, convolutional neural networks (CNNs)~\cite{tsinalis2016automatic2,vilamala2017deep,chambon2018deep,olesen2018deep,sors2018convolutional,eldele2021attention}, recurrent neural networks (RNNs)~\cite{michielli2019cascaded,you2022automatic}, transformers~\cite{phan2022sleeptransformer}, and combinations thereof~\cite{supratak2017deepsleepnet,phan2018joint,seo2020iitnet,mousavi2019sleepeegnet,guillot2021robustsleepnet,phan2020towards,phan2021xsleepnet}.
A general trend in the automatic sleep scoring domain is to apply increasingly complex deep learning architectures over time. 
Along with this trend, we observed that the latest advancements require non-trivial learning procedures to apply their solutions to new (smaller) datasets, for instance, fine-tuning~\cite{phan2020towards,guillot2021robustsleepnet,phan2022sleeptransformer,phan2022sleeptransformer} or complex training procedures~\cite{phan2021xsleepnet,supratak2020tinysleepnet}.

In contrast to this general trend, architectures like TinySleepNet~\cite{supratak2020tinysleepnet}, SimpleSleepNet~\cite{guillot2020dreem__simplesleepnet}, and DeepSleepNet-lite~\cite{fiorillo2021deepsleepnetlite} focussed on more lightweight deep learning solutions. The satisfactory performance of these approaches indicate that most of the existing deep learning architectures are overly complex, resulting in models that are data-hungry and computationally demanding.

A major drawback to these research efforts on deep learning is that clinical acceptance is directly hindered by the obstacles that deep learning introduces~\cite{fiorillo2019automated}. 
We argue that employing deep learning models in a medical context is challenging as they are hard to train, deploy, and reproduce, while lacking interpretability~\cite{amann2020explainability,fichman2011editorial}. 
Training a deep learning model is a complex task as, on the one hand, such models require a lot of data, and, on the other hand, training itself typically involves specialized hardware (GPUs), data augmentation, initialization \& regularization procedures, special scheduling to update the learning rate, and more.
Moreover, the lack of interpretability in deep learning models, categorized as black-box systems, is a common skepticism in healthcare and medicine~\cite{amann2020explainability}.
As a result, many efforts in literature focussed on solving these challenges with deep learning models by researching data-efficiency~\cite{phan2020towards,phan2021xsleepnet,supratak2020tinysleepnet}, interpretability~\cite{phan2022sleeptransformer,pathak2021stqs}, or model uncertainty~\cite{fiorillo2021deepsleepnetlite}.

\section{Approach}\label{sec:approach}
Our approach follows, just as the previous work mentioned in Section~\ref{sec:related_work}, the traditional flow of conventional machine learning pipelines, i.e., preprocessing, feature extraction, and modeling. 
Our pipeline takes two EEG, one EOG, and one EMG signal as input\footnote{We did not include ECG as this was not available in the public sleep scoring datasets.}.

\subsection{Preprocessing}
Preprocessing is concerned with cleaning or transforming the raw data, to retain the relevant signal whilst removing artifacts or noise.
To that end, the EEG and EOG signals are filtered to only keep frequencies between 0.40 Hz and 30 Hz. This band-pass range is clinically supported to capture the meaningful frequencies of sleep-wave patterns~\cite{malhotra2013performance}. 
The EMG signal is band-pass-filtered between 0.50 Hz and 10 Hz, in line with~\cite{guillot2020dreem__simplesleepnet,vallat2021open}. Note that these band-pass filters remove the powerline noise, which manifests itself at either 50 Hz or 60 Hz (depending on the local electrical grid specification).
No artifact removal was applied to the PSG data. For inter-dataset operability, the EEG signals were resampled to 100 Hz. On average, preprocessing takes less than 2 s for 12 h of PSG data.

\subsection{Feature extraction}
Feature extraction aims to extract a set of characteristics, i.e., features, with the intention of constructing an expressive (lower-dimensional) representation of the data. 
We calculate a set of 131 features per window, these features are multi-domain (extracted from time and frequency domain) and multi-resolution (calculated on multiple window sizes).
We utilize \textit{tsflex} to realize this strided-window feature extraction~\cite{vanderdonckt2021tsflex}.
Table~\ref{tab:feature-functions} lists the feature functions that are applied to the data. These feature functions originate from the \textit{YASA}~\cite{vallat2021open} (Yet Another Spindle Algorithm) and \textit{tsfresh}~\cite{christ2018tsfresh} toolkit, both these packages are integrated in \textit{tsflex}. In total, these feature functions output 131 values\footnote{Remember when calculating the number of features that our pipeline uses two EEG signals.}.

\begin{table}
\centering
\caption{The feature functions, consisting of both time-domain and frequency-domain functions.
The \textit{binned entropy} feature first bins the time series to then sum up the entropy of the bins (this feature is calculated four times for each time a different number of bins; 5, 10, 30, and 60 bins). The \textit{spectral Fourier statistics} feature calculates the spectral centroid (mean), variance, skew, and kurtosis of the absolute Fourier transform spectrum. The \textit{binned Fourier entropy} feature calculates the binned entropy of the power spectral density (this feature is calculated for seven different bin sizes, i.e., 2, 3, 5, 10, 30, 60, and 100 bins). 
The \textit{applied frequency bands} are the slow delta (0.4-1 Hz), fast delta (1-4 Hz), theta (4-8 Hz), alpha (8-12 Hz), sigma (12-16 Hz), and beta (16-30 Hz) band. The frequency-domain features are based on a Welch’s periodogram with a 5-s window (i.e., a 0.2 Hz resolution).}
\label{tab:feature-functions}
\begin{tabular}{lcccc}
\toprule
\textbf{Function} &  EEG & EOG & EMG & \textit{\# features}\\
\midrule
\textbf{Time-domain} \\
std, iqr, skewness, kurtosis &  \checkmark & \checkmark & \checkmark & 16 \\
number of zero-crossings &  \checkmark & \checkmark & \checkmark & 4 \\
Hjorth mobility, Hjorth complexity &  \checkmark & \checkmark & \checkmark & 8\\
higuch fractal dimension, petrosian fractal dimension &  \checkmark & \checkmark & \checkmark & 8 \\
permutation entropy, \textit{binned entropy (4)} & \checkmark & \checkmark & \checkmark & 20 \\
\midrule
\textbf{Frequency-domain}\\
\textit{spectral Fourier statistics (4)} & \checkmark & \checkmark & \checkmark & 16 \\
\textit{binned Fourier entropy (7)} & \checkmark & \checkmark & \checkmark & 28 \\
Absolute spectral power in the 0.4-30 Hz band  & \checkmark & \checkmark & & 3 \\
Relative spectral power in the \textit{applied frequency bands (6)} & \checkmark & \checkmark & & 18 \\
fast delta + theta spectral power  & \checkmark & & & 2 \\
alpha / theta spectral power & \checkmark & & & 2 \\
delta / beta spectral power & \checkmark & & & 2 \\
delta / sigma spectral power & \checkmark & & & 2 \\
delta / theta spectral power & \checkmark & & & 2 \\
\bottomrule
\end{tabular}
\end{table}

\begin{figure}[htb]
    \centering
    \includegraphics[width=\textwidth]{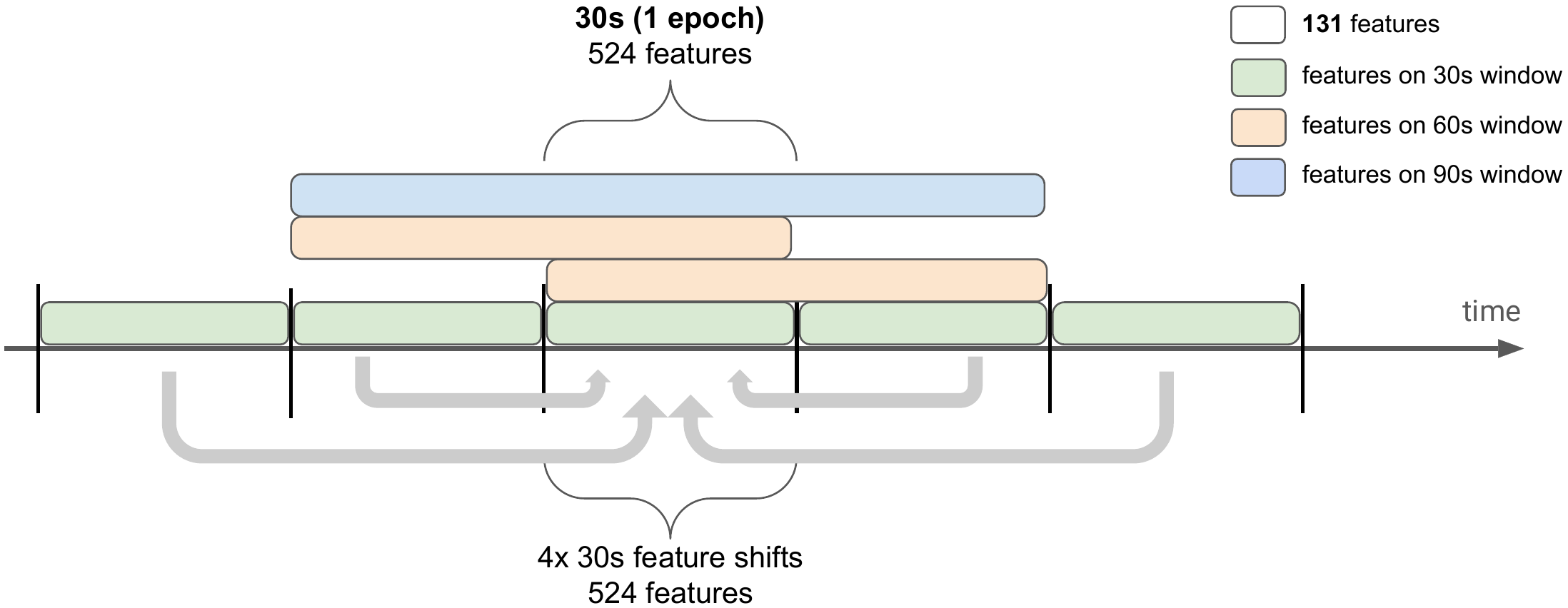}
    \caption{Multi-resolution and multi-domain feature extraction (for 1 epoch). Each box represents a collection of 131 multi-domain features. This same set of 131 features is extracted for three different window sizes (30 s, 60 s, and 90 s). The 90-s window is centered over the current epoch, while the two 60-s windows include the current epoch once at the start and once at the end of the window. 
    Additionally, features from the two preceding and two following 30-s windows are utilized for the epoch of interest.
    In total, the feature vector contains 1048 features.
    }
    \label{fig:feat_extract_windows}
\end{figure}

Figure~\ref{fig:feat_extract_windows} shows the various windows and shifts that are involved in creating the feature vector for an epoch\footnote{Note that including data from future epochs does not result in data leakage, as sleep stage classification is performed after the full recording was collected. Thus, at scoring time, the physician also has access to the full data (including future epochs).}.
The figure depicts how temporal context is included in the feature vector in two different ways.
On the one hand, the features are calculated over three different windows; i.e., 30 s, 60 s, and 90 s (each with a stride of 30 s). Such different window sizes result in other feature distributions, capturing a larger temporal range of the time series data. 
On the other hand, by incorporating the features that were extracted over 30-s windows from past and future windows, the feature vector contains fine-grained temporal context from up to two epochs before and after the current epoch. These shifted features allow differentiating the surrounding epochs from the current epoch, as the shifted features are directly comparable with the 30 s feature from the current epoch.
Calculating the features over the various windows, with four different resolutions, results in a feature vector of length 524 (see part above the time axis in Figure~\ref{fig:feat_extract_windows}). Additionally, shifting the features calculated on the 30-s windows two epochs forward and backwards, adds another 524 features to the vector (see part under the time axis in Figure~\ref{fig:feat_extract_windows}). Hence, the resulting feature vector has a dimensionality of 1048 (summarizing 75,000 data points). On average, feature extraction takes less than 20 s for 12 h of PSG data.

\subsection{Modeling}
We consider two different machine learning models, allowing us to assess the impact of model complexity on performance. In particular, we apply a linear model (logistic regression) and a non-linear model (gradient boosted trees).

The linear modeling is realized with the \textit{scikit-learn} toolkit~\cite{pedregosa2011scikit} and consists of two steps; (1) feature transformation, and (2) a linear machine learning model. 

In the first step, non-linear scaling is applied to transform the raw feature vector into a representation that is more suitable for the linear model. To that end, a quantile transformation (with 100 quantiles per feature) performs a monotonic operation to map the features to a uniform distribution, reducing the impact of outliers. This scaling method was selected as a considerable amount of the features distributions were heavily skewed.
As the second and final step, a logistic regression model is fitted on the transformed features.  

In contrast, the non-linear modeling has no additional feature transformation steps since tree-based models are less susceptible to skewed feature distributions. Thus, the non-linear modeling consists of only one component, i.e., a gradient boosted tree classifier. In this study, we used the CatBoost library~\cite{dorogush2018catboost} as this typically yields good classification results without intensive hyper-parameter tuning. Moreover, it has been shown that gradient boosted trees achieve state-of-the-art performance on many datasets~\cite{chen2016xgboost}.

Remark that we decided to not include feature selection in our proposed pipeline as (1) empirical results only showed marginal gains, (2) both models are rather robust to overfitting, and (3) including feature selection would not change the message of this paper\footnote{The first two arguments are supported by the experiments provided in the code repository in the \texttt{feature\_selection.ipynb} notebook.}. Moreover, no hyperparameter tuning was performed, given the little to no overfitting we observed in the learning curves\footnote{The learning curves can be found back in the experiment notebooks in the code repository}.

\section{Experiments}\label{sec:experiments}
In this section, we first describe the datasets that will be used in the study. Afterwards, the evaluation setup is detailed. Finally, we discuss the results and investigate the feature vector.

\subsection{Datasets}
Table~\ref{tab:datasets} shows an overview of the utilized datasets. For each dataset, we provide the distribution of sleep stages, accompanied by some metadata. The size of the employed datasets range from 40 to 153 night recordings. For more details on the collected data (e.g., subject statistics, inclusion and exclusion criteria), we further refer to the dataset description papers~\cite{sleepedf,o2014montrealmass}

\begin{table}[htp]
\centering
\caption{The distribution of sleep stages of each dataset. For \texttt{SC-EDF-20} and \texttt{SC-EDF-78}, the wake periods were trimmed to 30 minutes before and after the sleep period. The \textit{\# p} and \textit{\# r} columns refer respectively to the number of patients and recordings.}
\label{tab:datasets}
\resizebox{\textwidth}{!}{
\begin{tabular}{p{0.18\textwidth}p{0.07\textwidth}p{0.07\textwidth}p{0.135\textwidth}p{0.15\textwidth}p{0.08\textwidth}p{0.08\textwidth}p{0.08\textwidth}p{0.08\textwidth}p{0.11\textwidth}p{0.08\textwidth}}
\toprule
\multicolumn{1}{c}{\multirow{2}{*}{\textbf{Dataset}}} & \multicolumn{4}{c}{Metadata} & \multicolumn{6}{c}{Sleep stages (\# epochs)} \\
\cmidrule(lr){2-5} \cmidrule(lr){6-11} 
\multicolumn{1}{c}{} & \textbf{\# p} & \textbf{\# r} & \textbf{protocol} & \textbf{age $\pm$ std} & \textbf{W} & \textbf{N1} & \textbf{N2} & \textbf{N3} & \textbf{REM} & \textbf{Total} \\ 
\midrule 
\textbf{SC-EDF-20} & 20 & 40 & R \& K & 28.7 $\pm$ 2.9 & 8207 & 2804 & 17799 & 5703 & 7717 & 42230 \\
\textbf{SC-EDF-78} & 78 & 153 & R \& K & 58.8 $\pm$ 22.0 & 65642 & 21520 & 69132 & 13039 & 25835 & 195168 \\
\textbf{ST-EDF} & 22 & 44 & R \& K & 40.2 $\pm$ 17.7 & 4488 & 3653 & 19851 & 6415 & 8349 & 42756 \\
\textbf{MASS SS3} & 62 & 62 & AASM & 42.5 $\pm$ 18.9 & 6442 & 4839 & 29802 & 7653 & 10581 & 59317 \\
\bottomrule
\end{tabular}
}
\end{table}

\subsubsection{Sleep-EDF Database Expanded} The 2018 \texttt{Sleep-EDF} dataset~\cite{sleepedf} consists of two subsets; \textit{Sleep Cassette} (SC) and \textit{Sleep Telemetry} (ST). 

The Sleep Cassette subset contains 153 PSG recordings belonging to 78 subjects. For all patients, except three, recordings of the first and the second night are available. Each recording contains the following signals of interest; 2 EEG (Fpz-CZ and Pz-Cz), 1 EOG (horizontal), and 1 EMG (submental chin) signal. The PSGs also contain oro-nasal respiration and rectal body temperature. The EMG and EOG channels were sampled at 100 Hz. The EMG signal was electronically high-pass filtered, rectified and low-pass filtered, after which the RMS value (root-mean-square) was sampled at 1Hz\footnote{Note that this preprocessing is conflicting with the band-pass filter that our pipeline applies to the EMG signal. But, since the authors published only the preprocessed dataset, we have no choice but to accept this preprocessing.}.

We consider two splits, \texttt{SC-EDF-20} with subjects 0 to 19 from the SC study and \texttt{SC-EDF-78} with all the subjects from the SC subset\footnote{\texttt{SC-EDF-20} and \texttt{SC-EDF-78} refer respectively to the 2013 and 2018 version of the SC Sleep-EDF cohort. \texttt{SC-EDF-20} and \texttt{SC-EDF-78} are also often referred to as respectively \texttt{Sleep-EDF-v1} and \texttt{Sleep-EDF-v2}.}.
As the PSG recordings contain a lot of (wake) data before and after the sleep period, we only consider the PSG data between 30 minutes before and after the sleep period. This is the same protocol as other work~\cite{supratak2017deepsleepnet,perslev2021usleep,phan2021xsleepnet,phan2022sleeptransformer,supratak2020tinysleepnet,guillot2021robustsleepnet,mousavi2019sleepeegnet,seo2020iitnet} allowing a fair comparison, and has no impact on the feature distribution as there is no time information in the features.

The Sleep Telemetry subset contains 44 PSG recordings belonging to 22 patients. The goal of the ST study was to research the effect of temazepam, a drug used to treat insomnia, on sleep. For each patient, two nights were recorded, one of which was after temazepam intake, and the other of which was after placebo intake. The PSGs contain two EEG (Fpz-Cz and Pz-Oz), one EOG (horizontal), and one EMG (submental chin) signals, all sampled at 100 Hz.

Both subsets are scored according to the Rechtschaffen and Kales (R\&K) rules~\cite{rechtschaffen1968manual}. We convert the sleep stage labels to the AASM standard by merging N3 and N4 into a single N3 stage. To facilitate comparison with other work~\cite{guillot2021robustsleepnet,phan2021xsleepnet,supratak2020tinysleepnet,guillot2020dreem__simplesleepnet,mousavi2019sleepeegnet,phan2020towards,seo2020iitnet,supratak2017deepsleepnet,phan2022sleeptransformer,fiorillo2021deepsleepnetlite}, epochs labeled as \textit{MOVEMENT} or \textit{UNKNOWN} are excluded.

\subsubsection{MASS}
The Montreal Archive of Sleep Studies (MASS) dataset~\cite{o2014montrealmass} consists of five subsets (SS1 - SS5). In this study, we consider the \texttt{MASS SS3} cohort, which is composed of 62 nights from healthy subjects. Each recording contains 20 scalp-EEG, 2 EOG, 3 EMG, and 1 ECG channels. All EOG and EEG signals have a sampling rate of 256 Hz, whereas the EMG channels were sampled at either 128 Hz (in 43 recordings) or 256 Hz (in 19 recordings). 
Manual annotation was performed by sleep experts according to the AASM standard~\cite{berry2012rules}.

We consider two EEG (F4-EOG (Left) and F8-Cz), one EOG (average of left and right EOG), and one EMG (average of EMG1 and EMG2) signal. This selection was preferred as (i) it results in the same number of utilized EEG, EOG, and EMG channels as for the Sleep-EDF dataset, and (ii) is similar to the selection of other work~\cite{guillot2020dreem__simplesleepnet,supratak2020tinysleepnet,supratak2017deepsleepnet,seo2020iitnet}. All signals are downsampled to 100 Hz, improving the computational efficiency of the feature extraction.

\subsection{Evaluation setup}
In literature, various works have considered different training procedures. Among these procedures are Learning From Scratch (LFS), and Fine-Tuning (FT), and Direct Transfer (DT)~\cite{guillot2021robustsleepnet}: 

\begin{enumerate}
    \item \textit{LFS}: The model is trained from scratch on the current dataset and evaluated with a cross-validation procedure on the evaluation dataset. 
    \item \textit{FT}: The model is (pre-)trained on another dataset and fine-tuned with a cross-validation procedure on the current dataset. 
    \item \textit{DT}: The model is trained on another dataset and is evaluated (without fine-tuning) on the current dataset.
\end{enumerate}

It is essential to use the patient identifier as group in the cross validation procedures (when employing LFS and FT). This ensures that data from each patient was never in both the training and test fold (in the same iteration), preventing data leakage.
In this work, our proposed pipelines were evaluated according to the LFS procedure with $k$-fold cross-validation (CV). For SC-EDF-20 and ST-EDF, we evaluated using a $k$ of respectively 20 and 22 (in each fold one patient was left out as test set). For SC-EDF-78 and MASS SS3 we considered 10 folds. These configurations allow comparison with many other works~\cite{guillot2021robustsleepnet,phan2021xsleepnet,supratak2020tinysleepnet,guillot2020dreem__simplesleepnet,mousavi2019sleepeegnet,seo2020iitnet,supratak2017deepsleepnet,phan2022sleeptransformer}.
For SC-EDF-20 we also evaluated our pipeline via \textit{direct transfer}. To that end, we trained our pipeline on the data from the 58 patients of Sleep-EDF-78 that are not part of SC-EDF-20.

\subsection{Results}

Table~\ref{tab:results} presents a comparison of our pipeline to other (deep learning) work\footnote{Note however that we cannot facilitate a statistical comparison as most other works do not provide individual results per subject}. 
The results in the table indicate that we achieve state-of-the-art performance on the two smaller datasets (\texttt{SC-EDF-20} and \texttt{ST-EDF-20}). Note that for both datasets only RobustSleepNet is able to achieve higher performance, but as that approach uses fine-tuning and thus leverages more data, the results are not really comparable and favoring RobustSleepNet. On the two larger datasets, our simple pipelines are competitive in performance when comparing them to the state-of-the-art deep learning models. Observe that, for our proposed approach, including EMG in addition to EEG+EOG results in a 0.001-0.008 improvement for all three metrics\footnote{Additional experiments investigating the impact of various input (subset) combinations on performance, can be found back in the code repository.}.


We observe that the non-linear model consistently results in higher performance compared to the linear model. This is an indication that the linear model, due to its high bias, underfits the data (to some minor degree). 
Furthermore, considerable differences between the ranking of metrics are noticeable; e.g., a higher macro F1 score does not necessarily translate to the best accuracy score and vice versa. Hence, this empirically shows that comparing models with a single metric is not desirable. Remark that none of the metrics include certainty (i.e., the probabilities of the model). We believe that a log loss score might be a fitting way to assess this\footnote{Log loss is not included in the comparison table, as this metric was mostly not reported in other comparable work.}.

\subsection{Feature vector analysis}

\begin{figure}[htbp]
    \centering
    \includegraphics[width=0.8\textwidth]{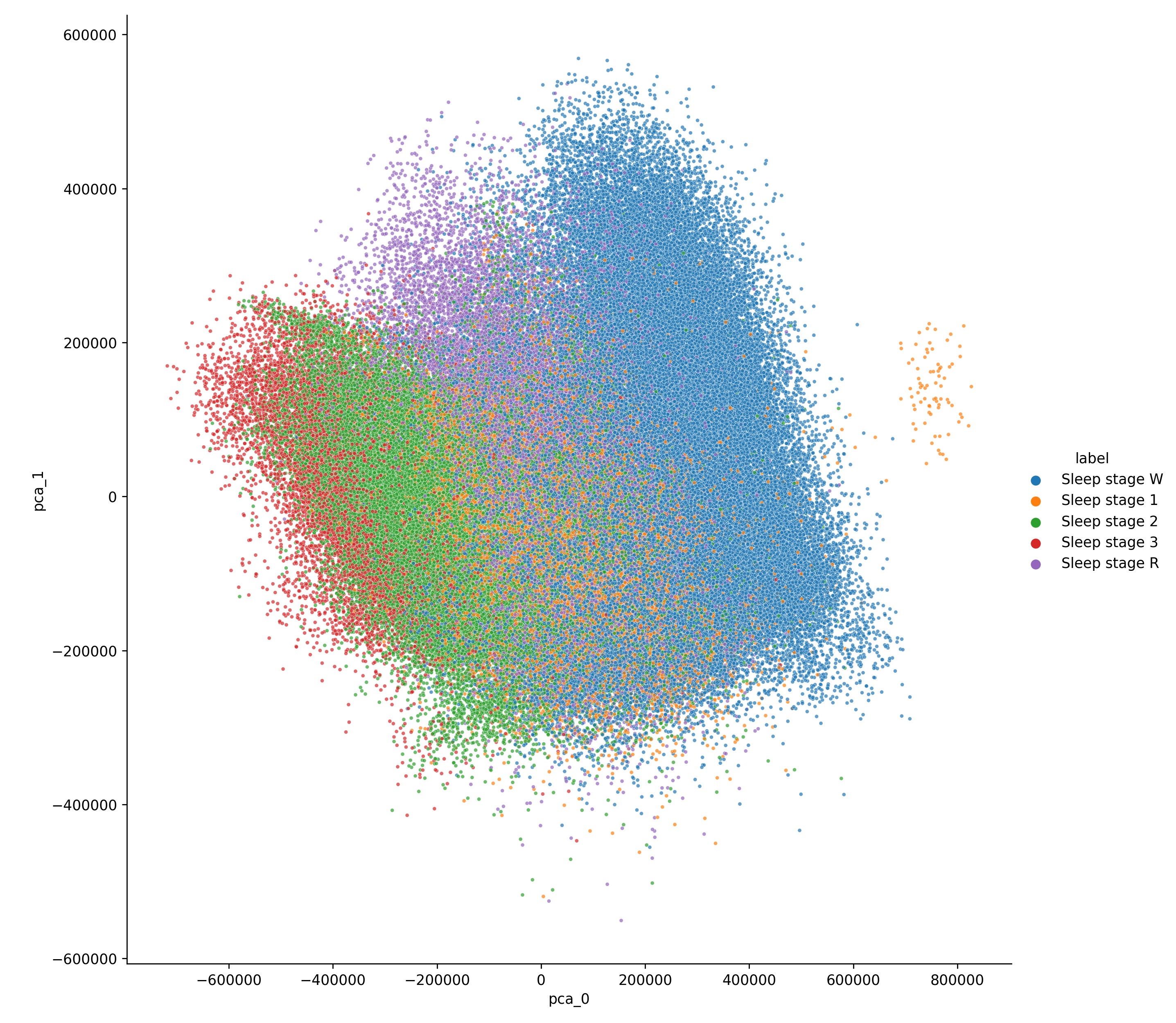}
    \caption{PCA projection in 2 components for all samples (i.e., feature vectors) from the \texttt{SC-EFD-78} dataset.}
    \label{fig:pca}
\end{figure}

Figure~\ref{fig:pca} shows a 2D projection of the feature vector when utiluizing the two components with highest explained variance from a principal component analysis (PCA) decomposition\footnote{In the code repository we also include a t-SNE projection of the feature vector. This projection is very similar to the PCA projection.}. Remark that PCA performs a linear transformation on the feature vector. It can be observed that this unsupervised dimensionality reduction technique allows a good separation of the sleep stages (except for \textsc{n1}), with just two PCA components that are a linear combination of the feature vector. On top of that, the distance between the clusters of various sleep stages in this projection space is rather intriguing. We observe that \textsc{rem} is closest to wake, as \textsc{rem} EEG frequencies and EOG patterns have close similarity to the wake state. \textsc{n3} is most distanced from wake, \textsc{n2} is between \textsc{n3} and \textsc{rem}, while \textsc{n1} is scattered all over the place. As such, it appears that the x-axis corresponds to the depth of sleep. 

\subsection{Code availability \& Reproducibility}
The code and results from this study are made available on GitHub under an open source license \url{https://github.com/predict-idlab/sleep-linear} to enable reproducibility of all results. In this repository you can also find more detailed results, including learning curves, confusion matrices, feature importances, and log loss scores.

\begin{table}[htbp]
\centering
\caption{Comparison between the proposed classical machine learning pipeline and other (deep learning) solutions using macro F1-score (MF1), overall accuracy (ACC), and Cohen's Kappa coefficient ($\kappa$). The approaches are sorted according to macro F1. The scores in bold represent the best score for each dataset (that are comparable to our approach).}
\label{tab:results}
\resizebox{\textwidth}{!}{
\begin{tabular}{c|p{0.07\textwidth}p{0.27\textwidth}p{0.2\textwidth}p{0.07\textwidth}|ccc|l}
\toprule
\multicolumn{1}{l}{\textbf{Dataset}} & \textbf{Year} & \textbf{System} & \textbf{Technique} & \textbf{LP} & \textbf{MF1} & \textbf{ACC} & \textbf{$\kappa$} & \textbf{Signals} \\
\midrule
\multirow{19}{*}{\textbf{Sleep-EDF-SC-20}} & 2021 & RobustSleepNet~\cite{guillot2021robustsleepnet} & RNN & FT & 0,817 & - & - & EEG + EOG \\
 & 2022 & \textit{This work} & Catboost & DT & \textbf{0,810} & \textbf{0,866} & \textbf{0,816} & EEG + EOG + EMG \\
 & 2021 & XSleepnet2~\cite{phan2021xsleepnet} & CNN \& RNN & LFS & 0,809 & 0,864 & 0,813 & EEG + EOG \\
 & 2022 & \textit{This work} & Logistic regr. & LFS & 0,809 & 0,857 & 0,806 & EEG + EOG + EMG \\
 & 2022 & \textit{This work} & Logistic regr. & DT & 0,805 & 0,863 & 0,813 & EEG + EOG + EMG \\
 & 2020 & TinySleepNet~\cite{supratak2020tinysleepnet} & CNN \& RNN & LFS & 0,805 & 0,854 & 0,800 & EEG \\
 & 2020 & SimpleSleepNet~\cite{guillot2020dreem__simplesleepnet} & RNN & LFS & 0,805 & - & - & EEG + EOG \\
 & 2022 & \textit{This work} & Logistic regr. & LFS & 0,803 & 0,853 & 0,800 & EEG + EOG \\
 & 2022 & \textit{This work} & Catboost & LFS & 0,802 & 0,864 & 0,812 & EEG + EOG + EMG \\
 & 2020 & XSleepnet1~\cite{phan2021xsleepnet} & CNN \& RNN & LFS & 0,798 & 0,852 & 0,798 & EEG + EOG \\
 & 2022 & \textit{This work} & Catboost & LFS & 0,797 & 0,860 & 0,807 & EEG + EOG \\
 & 2019 & SleepEEGNet~\cite{mousavi2019sleepeegnet} & CNN \& RNN & LFS & 0,797 & 0,843 & 0,790 & EEG \\
 & 2020 & SeqSleepNet+~\cite{phan2020towards} & RNN & FT & 0,796 & 0,852 & 0,789 & EEG \\
 & 2021 & RobustSleepNet~\cite{guillot2021robustsleepnet} & RNN & LFS & 0,791 & - & - & EEG + EOG \\
 & 2021 & RobustSleepNet~\cite{guillot2021robustsleepnet} & RNN & DT & 0,791 & - & - & EEG + EOG \\
 & 2020 & DeepSleepNet+~\cite{phan2020towards} & CNN & FT & 0,790 & 0,846 & 0,782 & EEG + EOG \\
 & 2021 & DeepSleepNet-Lite~\cite{fiorillo2021deepsleepnetlite} & CNN & LFS & 0,780 & 0,840 & 0,780 & EEG \\
 & 2019 & IITNet~\cite{seo2020iitnet} & CNN \& RNN & LFS & 0,776 & 0,839 & 0,780 & EEG \\
 & 2017 & DeepSleepNet~\cite{supratak2017deepsleepnet} & CNN \& RNN & FT & 0,769 & 0,820 & 0,760 & EEG \\
 \midrule
\multirow{14}{*}{\textbf{Sleep-EDF-SC-78}} 
 & 2022 & SleepTransformer~\cite{phan2022sleeptransformer} & transformer & FT & 0,788 & 0,849 & 0,789 & EEG \\
 & 2021 & XSleepnet2~\cite{phan2021xsleepnet} & CNN \& RNN & LFS & \textbf{0,787} & \textbf{0,840} & \textbf{0,778} & EEG + EOG \\
 & 2020 & XSleepnet1~\cite{phan2021xsleepnet} & CNN \& RNN & LFS & 0,784 & 0,840 & 0,777 & EEG \\
 & 2020 & TinySleepNet~\cite{supratak2020tinysleepnet} & CNN \& RNN & LFS & 0,781 & 0,831 & 0,770 & EEG \\
 & 2021 & RobustSleepNet~\cite{guillot2021robustsleepnet} & RNN & FT & 0,779 & - & - & EEG + EOG \\
 & 2022 & \textit{This work} & Catboost & LFS & 0,775 & 0,831 & 0,766 & EEG + EOG + EMG \\
 & 2022 & \textit{This work} & Catboost & LFS & 0,772 & 0,830 & 0,763 & EEG + EOG \\
 & 2022 & \textit{This work} & Logistic regr. & LFS & 0,771 & 0,821 & 0,756 & EEG + EOG + EMG \\
 & 2022 & \textit{This work} & Logistic regr. & LFS & 0,768 & 0,820 & 0,753 & EEG + EOG \\
 & 2021 & RobustSleepNet~\cite{guillot2021robustsleepnet} & RNN & LFS & 0,763 & - & - & EEG + EOG \\
 & 2021 & DeepSleepNet-Lite~\cite{fiorillo2021deepsleepnetlite} & CNN & LFS & 0,752 & 0,803 & 0,730 & EEG \\
 & 2022 & SleepTransformer~\cite{phan2022sleeptransformer} & transformer & LFS & 0,743 & 0,814 & 0,743 & EEG \\
 & 2021 & RobustSleepNet~\cite{guillot2021robustsleepnet} & RNN & DT & 0,738 & - & - & EEG + EOG \\
 & 2019 & SleepEEGNet~\cite{mousavi2019sleepeegnet} & CNN \& RNN & LFS & 0,736 & 0,800 & 0,730 & EEG \\
 \midrule
\multirow{9}{*}{\textbf{Sleep-EDF-ST}} & 2021 & RobustSleepNet~\cite{guillot2021robustsleepnet} & RNN & FT & 0,810 & - & - & EEG + EOG \\
 & 2022 & \textit{This work} & Catboost & LFS & \textbf{0,795} & \textbf{0,836} & \textbf{0,765} & EEG + EOG + EMG \\
 & 2022 & \textit{This work} & Logistic regr. & LFS & 0,792 & 0,829 & 0,759 & EEG + EOG + EMG \\
 & 2021 & RobustSleepNet~\cite{guillot2021robustsleepnet} & RNN & DT & 0,791 & - & - & EEG + EOG \\
 & 2022 & \textit{This work} & Catboost & LFS & 0,789 & 0,832 & 0,758 & EEG + EOG \\
 & 2022 & \textit{This work} & Logistic regr. & LFS & 0,788 & 0,825 & 0,754 & EEG + EOG \\
 & 2021 & RobustSleepNet~\cite{guillot2021robustsleepnet} & RNN & LFS & 0,786 & - & - & EEG + EOG \\
 & 2020 & DeepSleepNet+~\cite{phan2020towards} & CNN & FT & 0,775 & 0,815 & 0,738 & EEG \\
 & 2020 & SeqSleepNet+~\cite{phan2020towards} & RNN & FT & 0,775 & 0,810 & 0,734 & EEG \\
 \midrule
\multirow{12}{*}{\textbf{MASS SS3}} & 2020 & SimpleSleepNet~\cite{guillot2020dreem__simplesleepnet} & RNN & LFS & \textbf{0,847} & - & - & EEG + EOG \\
 & 2021 & RobustSleepNet~\cite{guillot2021robustsleepnet} & RNN & FT & 0,840 & - & - & EEG + EOG \\
 & 2020 & TinySleepNet~\cite{supratak2020tinysleepnet} & CNN \& RNN & LFS & 0,832 & \textbf{0,875} & \textbf{0,820} & EEG \\
 & 2021 & RobustSleepNet~\cite{guillot2021robustsleepnet} & RNN & LFS & 0,822 & - & - & EEG + EOG \\
 & 2022 & \textit{This work} & Catboost & LFS & 0,817 & 0,867 & 0,803 & EEG + EOG + EMG \\
 & 2017 & DeepSleepNet~\cite{supratak2017deepsleepnet} & CNN \& RNN & FT & 0,817 & 0,862 & 0,800 & EEG \\
 & 2022 & \textit{This work} & Catboost & LFS & 0,809 & 0,863 & 0,797 & EEG + EOG \\
 & 2021 & RobustSleepNet~\cite{guillot2021robustsleepnet} & RNN & DT & 0,808 & - & - & EEG + EOG \\
 & 2022 & \textit{This work} & Logistic regr. & LFS & 0,807 & 0,853 & 0,786 & EEG + EOG + EMG \\
 & 2019 & IITNet~\cite{seo2020iitnet} & CNN \& RNN & LFS & 0,805 & 0,863 & 0,790 & EEG \\
 & 2021 & U-Sleep~\cite{perslev2021usleep} & CNN & DT & 0,800 & - & - & EEG + EOG \\
 & 2022 & \textit{This work} & Logistic regr. & LFS & 0,794 & 0,845 & 0,775 & EEG + EOG \\
 \bottomrule
\end{tabular}
}
\end{table}

\section{Discussion}\label{sec:discussion}
In this section, we will discuss why our simple models match state-of-the-art predictive performances and what these results could imply for future research.

\subsection{A good feature vector is all you need}
Machine learning often boils down to transforming your data in a representation that is more suitable for prediction~\cite{domingos2012few}.
In particular, for classification tasks (such as sleep scoring), inference translates to separating the classes on the basis of a representation, i.e., creating decision boundaries within your data-representation space.

Deep learning is a popular machine learning approach nowadays as the task of transforming your data into a representation is learned by the model (the hidden layers of the model perform this transformation) instead of being constructed in a prior step. As a result, raw (scaled) data can be fed directly. 
In contrast, most classical machine learning models, such as linear models and tree-based models, do not learn internal representations on your raw data. Such models rely on learning relations on the supplied representations of the data that is fed to the model, e.g, linear relations when using a linear model or splitting thresholds when using decision trees.  
As a result, such models require an expressive feature vector as model input to work well.

We argue that classical machine learning is capable of achieving (near) state-of-the-art performance when using a feature vector that is highly representative for the sleep scoring task. This statement is confirmed by the impressive results when using a logistic regression classifier (see Table~\ref{tab:results}), as a linear model has limited expressiveness by learning linear relations on the input feature vector.

We believe that our feature vector exhibits several interesting properties, making it representative for sleep scoring as it is a (1) multi-resolution and (2) multi-domain summary of the multimodal PSG data, that (3) includes temporal context of the surrounding 2 epochs. 
The first two aspects are in line with the work of Nguyen et al.~\cite{le2019interpretable_ts_linear}, where the authors showed that a multi-resolution multi-domain linear classifier achieves similar accuracy as state-of-the-art (deep learning) methods. However, the authors focus on sequence classifiers, whereas our work focuses on more simple and interpretable classical machine learning models. The second aspect is also in line with Phan et al. their XSleepNet paper~\cite{phan2021xsleepnet}, as the proposed architecture's multi-view concept (using both raw signal data and time-frequency data) translates to multi-domain features in classical machine learning.
The third aspect, i.e., shifting the 30s-window features, adds more temporal context in the feature vector. As a result, the model itself does not need to handle the temporal relationship that is present in sleep stages as we embed this in the features. This is in contrast to the many works that employ sequence-to-sequence deep learning models~\cite{phan2020towards}.
Furthermore, time series feature extraction conveniently handles multimodal data, as the features from various modalities are simply concatenated in one feature vector.

As mentioned above, the PCA projection further illustrates the representativity of our feature vector.
These observations are in line with the results of Decat et al. where a feature-based cluster analysis was performed~\cite{decat2022beyond}, highlighting that the clusters substantially overlap with visual sleep scoring.

Finally, we want to stress that no new features were invented, nor were specific optimizations applied to the feature vector. All features were imported from existing libraries containing field-tested features (\textit{tsfresh}~\cite{christ2018tsfresh} and \textit{YASA}~\cite{vallat2021open}) and the strided-window feature extraction was conveniently realized with \textit{tsflex}~\cite{vanderdonckt2021tsflex}\footnote{We believe that recent advancements in the open source Python landscape enabled convenient and efficient creation of this feature vector. Especially the multi-window feature-extraction from \textit{tsflex}~\cite{vanderdonckt2021tsflex} allowed efficient creation of multi-resolution features, and \textit{plotly-resampler}~\cite{vanderdonckt2022plotly_resampler} enabled effective visual analysis of the data.}.
As such, we can argue that limited effort was required in constructing the feature vector. Thus, we believe that this work is first to provide a counterexample to the claims of several works stating that feature extraction is a cumbersome and time-intensive process~\cite{mousavi2019sleepeegnet,supratak2017deepsleepnet,seo2020iitnet}. Therefore, we hope that this work will serve as a strong feature-based baseline in future research with deep learning.

\subsection{Do not sleep on traditional machine learning}
Given the main focus in the automatic sleep scoring domain on deep learning today, \textit{we believe that a lot of researchers have been sleeping on traditional machine learning}. In other words, the impact of a representative feature vector and a simple machine learning model has been underestimated in this domain. Especially given the many advantages that this approach has over deep learning for clinical acceptance.

Section~\ref{sec:related_work_DL} highlights drawbacks of deep learning that are not or less prevalent in classical machine learning. To summarize, deep learning models (i) are hard to train, deploy and reproduce, and (ii) lack interpretability which results in black-box skepticism.
In contrast, although classical machine learning models are not fully white-box, they are arguably more interpretable than deep learning models. Moreover, the first limitation, i.e., deep learning models requiring a lot of data to properly generalize, is observable in our results, as for the smaller datasets we outperform all deep learning approaches (when learning from scratch). 
Furthermore, contrasting to deep learning models being both resource and time expensive to train, our linear models contain 1,000 to 10,000x less parameters and inference (including processing and feature extraction) takes under 25 sec for a typical night in \texttt{MASS SS3}. The linear model fits in under a minute on the largest dataset in this study (on a low-end CPU - Intel Xeon E5-2650 v2)\footnote{This training time does not include the time to preprocess and extract features, which is $\pm$ 20 sec for a typical night in MASS SS3. Note however that these steps should only happen once, as the extracted features from the processed data can be stored.}.

Our final remark is concerned with the minimal performance gains of the latest, complex (deep learning) solutions. Visual sleep scoring is inherently limited by a significant inter-scorer and intra-scorer disagreement~\cite{rosenberg2013american}, raising questions on the added value of those marginal improvements.

Considering all factors above, we believe that there is a strong case for paying more attention to conventional machine learning pipelines, consisting of simple models. Especially, the presented results show that using a more classical approach should not always come at the cost of performance.

\section{Conclusion}\label{sec:conclusion}
In contrast to the numerous deep learning approaches in literature, this work investigates a more classical approach for automatic sleep scoring.
In particular, we employ a conventional machine learning pipeline consisting of preprocessing, feature extraction, and a simple machine learning model.
Results show that our approach outperforms current state-of-the-art on two small datasets, while scoring competitively on two larger datasets. We argue that the strength of our pipeline lies in having a highly representative feature vector, which we demonstrate through a PCA projection and the performance of our linear model. Our feature vectors are a multi-resolution and multi-domain summary of the PSG data and include temporal context of surrounding epochs. Such an expressive feature vector enables more simple and widely accepted machine learning models. On top of that, our training times are merely a fraction of those of deep learning solutions.
Moreover, classical machine learning pipelines do not suffer from the challenges that are prevalent in deep learning, i.e., deep learning models being black-box models that are hard to train, deploy, reproduce, while lacking interpretability. These challenges themselves are directly hindering the broad clinical acceptance of deep learning models.

Given the strong performance of our simple pipeline together with the interpretability and successful track record in medicine of classical models, this work paves the path to adoption of classical sleep scoring algorithms in a clinical context. We further question if current research is targeting the right challenges by focusing on marginal improvements, certainly in the light of rather high inter-annotator disagreement.

With this work, we hope to raise a new perspective on automatic sleep scoring, where researchers are no longer sleeping on the performance of traditional machine learning.

\section*{Acknowledgements}
The authors thank Vic Degraeve and Jarne Verhaeghe for providing feedback on the manuscript.

%
%
%
\bibliographystyle{splncs04}
\bibliography{references.bib}

\end{document}